\documentclass[runningheads]{llncs}

\usepackage[T1]{fontenc}
\usepackage{graphicx}
\usepackage{hyperref}
\usepackage{color}
\usepackage{color}

\usepackage{amsmath}
\usepackage{amssymb}
\usepackage{bm}
\usepackage{xurl}
\usepackage[utf8]{inputenc}
\usepackage{multirow}
\usepackage{booktabs}
\usepackage{graphicx}
\usepackage{array}
\usepackage{subfig}
\usepackage{fancyhdr}
\begin{document}

\title{On Impact of Loss Function on the Performance of Neural Networks in Melanoma Diagnosis}
\titlerunning{Impact of Loss Function on Melanoma Classification}
\author{Morgan May\inst{1,4}\orcidID{0009-0006-1953-7600} \and
Pierpaolo Dondio\inst{2}\orcidID{0000-0001-7874-8762} \and
Simon Caton\inst{3}\orcidID{0000-0001-9379-3879}}
\authorrunning{M. May et al.}
\institute{Technological University Dublin, Dublin, Ireland \and
Dublin City University, Dublin, Ireland \and
University College Dublin, Dublin, Ireland \and
Research Ireland Centre for Research Training in Machine Learning, ML-Labs, Dublin, Ireland\\
\email{simon.caton@ucd.ie}}
\maketitle
\thispagestyle{fancy} 
\vspace{-0.2in}
\begin{abstract}
Melanoma is the deadliest type of skin cancer, whose early diagnosis is crucial for patients' survival.
Image classification using deep learning models has shown promising results for melanoma diagnosis.
However, the performance of these models on the melanoma datasets such as SIIM-ISIC melanoma classification dataset is a challenge due to the class imbalance.
One of the methods to deal with this challenge is using loss function modifications.
In this work, we have investigated the effect of different loss functions on the performance of deep neural networks.
We trained these networks using focal loss, logit-adjusted softmax cross-entropy (CE) loss, and weighted softmax CE loss, and we report different metrics for evaluating performance and uncertainty calibration. Our results suggest that focal loss delivers a good combination of performance in terms of AUC and uncertainty calibration in terms of expected calibration error (ECE) simultaneously.
\end{abstract}

\section{Introduction}
Skin cancer is a common skin disease whose incidence and mortality rates have been showing yearly increases. 
Early diagnosis and timely treatment of skin cancer, especially for its deadliest form, malignant melanoma, remain a critical
clinical issue globally \cite{wang2025recent}.
Recently, deep learning has revolutionised melanoma diagnosis by building highly accurate classification models \cite{cassidy2022analysis,esteva2017dermatologist,kaur2022melanoma,meedeniya2024skin,naseri2025diagnosis}.

Melanoma diagnosis is an imbalance classification task, where in the available clinical datasets, the number of positive samples (malignant) is outnumbered by the negative samples (benign), because of the data scarcity of malignant cases.  
Imbalanced data poses a challenge for generalization on minority classes and makes naive learning biased towards the majority classes \cite{menon2021longtail}.
The methods to cope with this challenge can be divided into four groups \cite{chen2024survey}: 
data-level, algorithmic-level, hybrid of both and ensemble learning.
One of the algorithmic level approaches is loss function modifications \cite{cao2019learning,cui2019class,lin2017focal,menon2021longtail}.

Authors in \cite{nazari2025going} compared the effect of focal loss \cite{lin2017focal}, weighted cross-entropy (CE) loss \cite{ronneberger2015u}, and large margin-aware focal (LMF) loss \cite{sadi2022lmfloss} on the performance of EfficientNet-B3 on melanoma classification. 
They report the results with careful hyperparameter tuning and show that LMF loss delivers higher AUC compared to focal loss and weighted CE. 
In this work, we investigate the effect of different loss functions, 
including focal loss \cite{lin2017focal}, 
logit adjusted softmax cross-entropy loss \cite{menon2021longtail}, 
and balanced softmax cross-entropy loss \cite{ren2020balanced}, 
on the performance of EfficientNet-B0, EfficientNet-B2, and EfficientNetV2-S 
by performing hyperparameter tuning using grid search, 
for each of the combinations of loss functions and models, and incorporating the evaluation of model uncertainty calibration.

\section{Methodology} \label{sec:methodology}
\vspace{-0.1in}
This work evaluates the effects of the choice of loss function on the performance of neural networks in melanoma classification task. 
As such, it does not intend to compare the models
against each other or maximize the performance of the models. 
Therefore, no additional steps such as ensembling is included.

\noindent \textbf{Dataset.}
We use the SIIM-ISIC 2020 Melanoma Classification dataset \cite{isic2020dataset}, which is
imbalanced with an imbalance ratio of about 55. 
We used the 33126 labeled images in the original train set for a train/validation/test split of 80/10/10.

\noindent \textbf{Models.}
We train two EfficientNet-B0 and EfficientNet-B2 \cite{tan2019efficientnet},
and EfficientNetV2-S \cite{tan2021efficientnetv2}, from an improved generation of EfficientNets, designed for better parameter efficiency and faster training compared to its predecessor, EfficientNet. 
The models are pretrained on ImageNet.

\noindent \textbf{Loss Functions.} 
We examine three loss functions that are used for imbalanced data: Balanced softmax cross-entropy (CE) \cite{ren2020balanced},
focal loss \cite{lin2017focal}, and logit-adjusted softmax cross-entropy loss \cite{menon2021longtail}.

\noindent \textbf{Performance metrics.}
We report AUC, F1-score, and recall for model performance evaluation, 
and expected calibration error (ECE) \cite{guo2017calibration,naeini2015obtaining} for the evaluation of confidence calibration \cite{guo2017calibration}.

\noindent \textbf{Hyperparameter tuning.}
For hyperparameter tuning, we perform grid search, 
including the learning rate for each model, tuned from $1e-4$ to $3e-4$. 
For each of the loss functions, we perform individual grid searches for the hyperparameters specific to each of the loss functions, 
such as logit-adjustment temperature for logit-adjusted softmax CE loss, chosen from $\{0.5, 0.8, 1.0, 1.1, 1.2, 1.5, 2.0\}$, 
and alpha (the balancing parameter) and gamma (the focusing parameter) for focal loss, chosen from $\{0.25, 0.5, 0.75, 0.98\}$ and $\{1.0, 2.0, 4.0\}$, respectively.
Balanced softmax CE does not have any specific hyperparameters to tune.

\section{Results and Discussion} \label{sec:eval}
\tablename~\ref{tab:summary} summarizes the results of training the models using different loss functions on SIIM-ISIC 2020 melanoma classification task. 
Initial results do not suggest a strong trend across all models trained using different loss functions. 
However, some insights can be extracted on the performance of the models trained using different loss functions.

For the bigger models, EfficientNet-B2 and EfficientNetV2-S, focal loss results in a higher AUC value compared to logit-adjusted softmax cross-entropy loss and balanced softmax cross-entropy loss, while maintaining a relatively low ECE value across all three models.
This suggests that focal loss is a good choice, specially for training bigger models to have a good performance, while maintaining good confidence calibration, 
which is crucial for a safety critical application such as melanoma diagnosis.
It can also be observed that for both EfficentNet-B0 and EfficientNet-B2, logit-adjusted softmax cross-entropy loss leads to a higher recall than focal loss and balanced softmax cross-entropy loss, while sacrificing the ECE value, which may result in overconfident predictions, which is not desirable for melanoma diagnosis as a safety-critical application.

\begin{table}[h]
    \vspace{-0.75cm}
    \centering
    \caption{Performance of different models on melanoma classification trained using different loss functions}
    \begin{tabular}{|l|l|c|c|c|c|}
        \hline
        \multicolumn{1}{|c|}{MODEL} & \multicolumn{1}{c|}{LOSS} & AUC & F1 & Recall & ECE \\ \hline
        \multirow{3}{*}{EfficientNet-B0} & Focal Loss & 0.8804 & \textbf{0.581} & 0.578 & \textbf{0.0142} \\ \cline{2-6} 
         & Logit Adjusted Softmax CE & 0.8840 & 0.532 & \textbf{0.740} & 0.0390 \\ \cline{2-6} 
         & Balanced Softmax CE & \textbf{0.8970} & 0.573 & 0.564 & 0.0212 \\ \hline \hline
        \multirow{3}{*}{EfficientNet-B2} & Focal Loss & \textbf{0.8909} & 0.581 & 0.585 & 0.0225 \\ \cline{2-6}
         & Logit Adjusted Softmax CE & 0.8904 & 0.405 & \textbf{0.788} & 0.0913 \\ \cline{2-6} 
         & Balanced Softmax CE & 0.8803 & \textbf{0.589} & 0.624 & \textbf{0.0206} \\ \hline \hline
        \multirow{3}{*}{EfficientNetV2-S} & Focal Loss & \textbf{0.9097} & 0.555 & 0.534 & \textbf{0.0163} \\ \cline{2-6} 
         & Logit Adjusted Softmax CE & 0.9027 & \textbf{0.626} & 0.623 & 0.0187 \\ \cline{2-6} 
         & Balanced Softmax CE & 0.8822 & 0.620 & \textbf{0.668} & 0.0213 \\ \hline
    \end{tabular}
    \label{tab:summary}
    \vspace{-1cm}
\end{table}

\section{Conclusion} \label{sec:conclusion}
This work evaluates the effect of different loss functions on the performance of imbalanced melanoma classification. The results suggest that focal loss is a good choice for training EfficientNets and EfficientNetV2 for melanoma classification, 
for achieving a good performance in terms of AUC, while being well-calibrated in terms of ECE, which is crucial for the safety critical application of melanoma diagnosis.
It will be interesting to see the effect of a wider variety of loss functions on the performance of a broader range of models for melanoma classification, 
and to evaluate the effect of a combination of multiple loss functions as the objective function for training the models.

\section{Acknowledgment}
This publication has emanated from research conducted with the financial support of Taighde Éireann – Research Ireland under Grant number 18/CRT/6183. 
For the purpose of Open Access, the author has applied a CC BY public copyright license to any Author Accepted Manuscript  arising from this submission. 

\bibliographystyle{splncs04}
\bibliography{Bibliography}

\end{document}